\documentclass{ieeeaccess}
\usepackage{cite}
\usepackage{amsmath,amssymb,amsfonts}
\usepackage{algorithmic}
\usepackage{graphicx}
\usepackage{textcomp}

\usepackage{balance}
\usepackage{multirow}%z
\usepackage{array} %Z
\usepackage{booktabs} %z

\def\BibTeX{{\rm B\kern-.05em{\sc i\kern-.025em b}\kern-.08em
    T\kern-.1667em\lower.7ex\hbox{E}\kern-.125emX}}
\begin{document}
\history{Date of publication xxxx 00, 0000, date of current version xxxx 00, 0000.}
\doi{10.1109/ACCESS.2023.0322000}

\title{Intelligent Detection of Mechanical, Electrical, and Plumbing (MEP) Metrics Based on 2D Floor Plans}
\author{
	\uppercase{Tarandeep Singh Mandhiratta}\authorrefmark{1}, %\IEEEmembership{Fellow, IEEE},
\uppercase{ANK Zaman}\authorrefmark{2}, \IEEEmembership{Member, IEEE} and \uppercase{Abdul-Rahman Mawlood-Yunis}\authorrefmark{3}, \IEEEmembership{Member, IEEE}}

\address[1]{Dept. of Physics \& Computer Science, Wilfrid Laurier University (WLU), Waterloo, ON, Canada (mand4710@mylaurier.ca)}
\address[2]{Dept. of Physics \& Computer Science, Wilfrid Laurier University (WLU), Waterloo, ON, Canada (e-mail: azaman@wlu.ca)}
\address[3]{Dept. of Physics \& Computer Science, Wilfrid Laurier University (WLU), Waterloo, ON, Canada (e-mail: amawloodyunis@wlu.ca)}

%\tfootnote{This paragraph of the first footnote will contain support information, including sponsor and financial support acknowledgment. For example, ``This work was supported in part by the U.S. Department of Commerce under Grant BS123456.''}

\markboth
{Author \headeretal: Preparation of Papers for IEEE TRANSACTIONS and JOURNALS}
{Author \headeretal: Preparation of Papers for IEEE TRANSACTIONS and JOURNALS}

\corresp{Corresponding author: ANK Zaman (e-mail: azaman@wlu.ca).}

\begin{abstract}
This research developed a neural network-based model to extract various information from 2D floor plans. We detect lighting symbols, identify the appropriate type of light, and extract the associated texts with lights. The study aims to enable efficient floor designing and determining the number and type of lights needed per floor, i.e., allow efficient design and estimate the power requirement of the floor plan. The model was developed using Mask RCNN as the base. The images were annotated and converted into a Coco data format for training the model. The model achieved bbox\_mAP and segm\_mAP values of 0.7596 and 0.7111, respectively. It also performed well at different IoU thresholds, i.e., with bbox\_mAP 50 and segm\_mAP 75 values of 0.9850 and 0.9219, respectively. The developed model will help various industries, such as architecture and construction, to improve design time and create efficient workflows by automatically detecting Mechanical, Electrical, and Plumbing (MEP) objects from floor plans, and it is the first step towards building tools that will help energy-efficient building design. 
\end{abstract}


\titlepgskip=-21pt

\maketitle

\section{Introduction}\label{sec:introduction}
\PARstart{T}{oday’s} technology has made collecting and processing information in indoor environments possible. This has positively impacted facilities management, indoor localization, and indoor data models, but it has also increased the demand for creating actual databases of indoor information.  

One area still underdeveloped ~\cite{WANG2022},~\cite{hu2023} is the analysis of the Mechanical, Electrical, and Plumbing (MEP) matrices of buildings, which is crucial for efficient engineering design processes. Various approaches have been used to analyze MEP matrices of buildings, including those that use 3D laser scanning, CAD plans, architectural sketches, and mobile applications. This study focuses on using 2D floor plans, as they provide the most fundamental source of information for existing buildings.  

Recent efforts have been put into using machine learning models for extracting and labelling data from floor plans. However, this is a challenging problem, as the notation of different floor plans varies significantly according to the institutions and the design offices. Furthermore, floor plans archived as complex, fuzzy architectural drawings usually characterize raster images.

This research work aims to intelligently collect MEP-related metrics from floor plans to aid in improving the efficiency of the engineering design process. As a first step, we developed a neural network model to collect three types of information from 2D floor plans: a) detect lighting symbols, b) identify the appropriate type of light, and c) extract the associated texts with lights. This work converted PDF files of 2D floor plans into image format to achieve these objectives. We annotated the images manually. This is because we use a private dataset provided by the sponsor, and the data is not annotated. Once the annotations were complete, the data was converted to coco data format. Each image and its coco dataset were sliced into smaller components to reduce storage space. The images were sliced with slice parameters such that each image was sliced into 1000x1000 grids with an overlap ratio of 0.2. After the data preprocessing step, Mask RCNN ~\cite{he2017},~\cite{matterport_maskrcnn_2017} was used as a base to develop a neural network model. The configuration files were changed to optimize the model for this use case, and hyperparameter tuning was also done. Initialization, forward propagation compute loss, and backpropagation were performed during training. The model was then trained and validated using a dataset to improve its accuracy. The model was used to detect the lighting symbols in the floor plan. Once the algorithm detects the lighting symbols, it finds the coordinates of the bounding boxes and extracts each electrical component detected, saving it in a file for use in the next step. The next step is to loop through the cropped lighting symbols saved in a file and extract the appropriate text that informs what kind of electrical component it is. Once the type of electrical component is identified, the power consumption for lighting components of the whole floor plan can be estimated by combining the results. Finally, TensorBoard ~\cite{tensorboard2019} was used to visualize the findings, which improved the model even more. 
This research represents a significant improvement over the traditional method of intelligently detecting MEP symbols from floor plans. The developed model will help various industries, such as architecture and construction, improve design time and create efficient workflows by automatically detecting Mechanical, Electrical, and Plumbing (MEP) objects from floor plans; it is the first step towards building tools that will help energy-efficient building design. 

The rest of the paper is organized as follows: section~\ref{sec2} surveys and presents recent related works, section~\ref{sec3} describes the technical details of the implementation and dataset used in the study, section~\ref{sec4} describes the result and evaluation, and section~\ref{sec5} describes the concluding remarks of this paper.
 
\section{Literature Review}\label{sec2}
\begin{figure*}[t!]%
	\centering
	\includegraphics[width=1.0\textwidth]{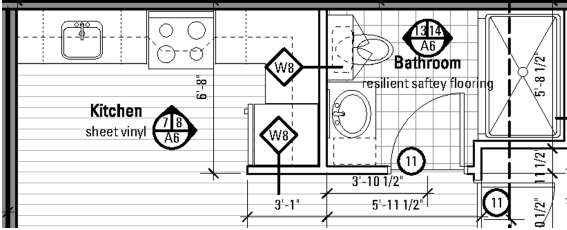}
	\caption{An architectural floor plan section, specifically a residential unit's kitchen and bathroom areas \cite{ rezvanifar2021analyzing}}\label{fig1}
\end{figure*}

In recent years, there has been a growing interest in using deep learning techniques to detect MEP (mechanical, electrical, and plumbing) metrics from 2D floor plans. Researchers have actively explored various approaches to extract components from architectural floor plans automatically. These floor plans, which architects create to provide a scaled representation of buildings, serve as a crucial source of information for MEP systems design and installation. These floor plans, created by architects, provide a scaled representation of the building and contain important details such as walls, doors, windows, stairs, fixtures, and labels that serve as critical information for MEP systems design and installation. For example, Figure~\ref{fig1} presents a floor plan for a residential unit's bathroom and kitchen areas \cite{ rezvanifar2021analyzing}.

Several studies have attempted to develop automated methods for detecting MEP systems from floor plans, utilizing machine learning algorithms such as Convolutional Neural Networks (CNNs) to classify symbols. The analysis of architectural floor plans has been a crucial area of research in computer vision. For instance, in 1992, Filipski \& Flanderena \cite{filipski1992automated} introduced GTX 5000, a commercially available tool that utilized neural networks for character and symbol detection, making it one of the first commercial applications of convolution image detection.

More recently, in 2017, Ruwanthika et al.~\cite{ruwanthika2017dynamic} attempted to construct complete 3D models of houses for AR purposes, while Dodge et al.~\cite{dodge2017parsing} applied Fully Convolution Networks (FCN) to analyze architectural floor plans with the goal of processing simplified plans frequently seen on real estate websites, which could not be effectively analyzed through conventional methods. The authors also compared traditional and statistical methods, utilizing Google Vision API to remove any text elements that could skew the results. %The comparison of these methods is shown in Figure~\ref{fig2}. 
Through their study, they demonstrated the precision with which one could identify the features of a floor plan. This breakthrough could serve as a building block for automated 3D modelling in the future. The renewed interest in neural networks for floor plan analysis can be attributed to significant advancements in deep learning and the availability of annotated data for training.

Automating the entire process of floor plan analysis without manual intervention is becoming more critical, especially with the increasing demand for smart homes and building management systems. As such, the following section provides an overview of the current techniques for extracting walls, detecting rooms, and performing preprocessing in this field, highlighting the difficulties encountered and the proposed solutions.

\subsection{Methods for Extraction of Walls}\label{sec2p1} 
The techniques used for wall extraction in architectural floor plans have advanced as the goals, difficulties, and technological advancements changed. The field has kept pace with the development of computer vision techniques and increased computing power.

\subsubsection{Early Techniques for Separating Walls in Floor Plans}\label{sec2p1p1} 
The first attempts to create systems for understanding architectural floor plans were driven by the need to convert large collections of drawings that were only available in printed or hand-drawn form into digital format. This task was often done manually, which could be time-consuming and expensive. It involved tracing the scanned image into CAD software either on a computer screen or using specialized tracing tables~\cite{ahsoon1997variations}. 

The beginning stages of solving the problem of digitizing architectural floor plans~\cite{ ahsoon1997variations},~\cite{aoki1996prototype},~\cite{koutamanis1992automated} only dealt with images with a set pattern of drawings, usually walls depicted as thick lines, and often were only able to handle horizontal and vertical lines. These early methods mainly focused on converting the walls into vectors using morphological operations and thinning the walls down to 1-pixel lines to obtain their vector representation. This traditional pipeline of line extraction, Skeletonization, and vectorization became a commonly used technique for analyzing technical drawings like topographic maps~\cite{greenlee1987raster}.

The earliest works aimed to convert floor plan sketches into CAD format efficiently. A study by Aoki et al. (1996)~\cite{aoki1996prototype} relied on the use of thin and thick lines to distinguish internal and external walls. The approach involved extracting line segments and closed regions from a preprocessed image and then using pattern matching to identify the elements in the drawing. 

Ahmed et al.~\cite{ahmed2012automatic} introduced an early automatic method for analyzing architectural floor plans that involved multiple stages, including information separation, structure analysis, and semantic analysis. However, their wall extraction technique was based on the assumption that walls were drawn as the thickest lines in the image, which restricted its use to that specific drawing style.

\subsubsection{Conventional Machine Learning Methods}\label{sec2p1p2}
\begin{figure}[h]%
	\centering
	\includegraphics[width=.45\textwidth]{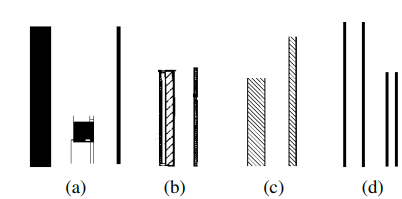}
	\caption{Various Wall Notations~\cite{6628813}}
	\label{fig:fig4}
\end{figure}

Heras et al.~\cite{heras2011wall} pointed out that architectural drawings have no universal notation. Walls can vary in representation, from a single line of varying widths to parallel lines or even hatching patterns, as shown in Figure~\ref{fig:fig4}. This diversity made traditional methods that rely on specific notations insufficient for wall extraction tasks.

They~\cite{heras2011wall} also introduced a groundbreaking machine-learning approach for wall extraction that can manage various notations, modelled on the traditional Bag-Of-Visual-Words model~\cite{ heras2013notation}. %The method is depicted in 

The authors in~\cite{heras2013unupervised} note the main limitations of patch-based methods as being restricted to single notation training and requiring ground truth data from a diverse image set for the training process.

\subsubsection{Methods Utilizing Convolution Neural Networks (CNN)}\label{sec2p1p3}

With the increasing computing power, CNNs have been widely adopted in computer vision and various fields, including architectural floor plan analysis. Liu et al.~\cite{ liu2017raster} used a simple CNN approach for heatmap prediction at the pixel level, as described in~\cite{bulat2016human}, to detect corners and crosspoints (referred to as "junctions") in floor plans. They trained their method using the LIFULL HOME dataset~\cite{ kiyota2018promoting}, which manually annotated 1,000-floor plan images. This dataset was large enough for deep learning. 

While the CNN-based approaches achieved high quantitative results, qualitatively, they failed to guarantee wall connectivity, sometimes missing entire walls, rendering post-processing recovery impossible.

\subsection{Techniques for Identifying Rooms}
Methods for detecting rooms focus on identifying and separating the various areas within a floor plan image, using information from the walls as a guide. Accuracy depends on correct and thorough wall extraction, as errors in wall identification can impact results.

\subsubsection{Pixel-level Approaches}
Pixel-based methods of room detection operate by analyzing a raster image of the walls. Koutamanis and Mitossi~\cite{koutamanis1992automated} proposed an early example of this method, which involved using fixed rules to fill gaps in a simplified version of the walls. However, this method was limited in its ability to handle complex cases or wall extraction errors. Another approach, proposed by Ryall et al.~\cite{ryall1995semi}, involved using a "proximity field" to identify the farthest points from any wall as potential room centers. Its reliance on pixel properties limited this method, resulting in potential  under-segmentation or over-segmentation of rooms.

\subsubsection{Geometry-Driven Approaches}
Geometry-based methods for room detection utilize a vector representation of the walls. Mace et al.~\cite{mace2010system} proposed an example of this method, which used a top-down polygon partitioning approach based on the idea that rooms should be nearly convex regions. This method filled gaps between walls by minimizing concavity in the resulting room polygon.

Other authors~\cite{ahsoon1997variations},~\cite{liu2017raster},~\cite{or2005highly},~\cite{heras2014statistical},~\cite{wessel2008room} have proposed methods for detecting loops in the polygon representation of walls, which are used to close gaps created by doors and windows. These methods often rely on prior detection of wall and symbol information with high precision, which is only feasible on datasets with low complexity or human assistance.

\subsubsection{Hybrid Approaches}
More advanced methods, such as the one proposed in~\cite{ahmed2011improved}, create a pixel image of the extracted walls and fill gaps in the wall structure using information from previous structural analysis steps.

Once the gaps are filled, the rooms are detected as connected components in the image. The hybrid methods utilize semantic and pixel information for room detection, balancing robustness and accuracy.

\subsection{Preprocessing Techniques}\label{sec2p3}

Graphics/text separation is a common preprocessing step in many wall extraction methods~\cite{ahsoon1997variations},~\cite{heras2011wall},~\cite{heras2013notation},~\cite{heras2013unupervised},~\cite{or2005highly} ,~\cite{heras2014statistical},~\cite{ahmed2011improved},~\cite{dosch2000complete}. This involves detecting and extracting text from the image to simplify it.

Fletcher and Kasturi~\cite{kasturi1990system} introduced a graphics/text separation technique in early wall extraction methods (Section~\ref{sec2p1}). The method relied on the distinct features of technical drawings to identify text characters. It filtered the image's connected components based on statistical histogram analysis of the component's area, aspect ratio, pixel density, and bounding box dimensions. This method was simple to implement and produced good results, but it could not detect text characters overlapping with other symbols.

Ahmed et al.~\cite{ ahmed2011text} improved the graphics/text separation method introduced by Fletcher and Kasturi~\cite{ kasturi1990system} by detecting overlapping text characters. The improved method inferred the location of missing text characters from surrounding detected characters. These methods relied on the assumption that there is sufficient text in the image to make the area of text characters the highest probability area measure among all connected components in the image.

One of the primary challenges in this area of research is the lack of a standardized symbol set for MEP systems. Different construction projects may use different symbols for the same MEP components. Developing a generalized model to detect all possible symbols for MEP systems across different projects is challenging.

Researchers have proposed several solutions to overcome this challenge, such as developing a symbol classification system that can learn from multiple symbol sets and using data augmentation techniques to generate synthetic data. Additionally, some researchers have proposed using an ensemble of models to increase the detection system's accuracy.

Overall, the research in this area shows great promise for developing intelligent detection systems for MEP metrics based on 2D floor plans. With the increasing availability of large-scale annotated data sets and advancements in deep learning techniques, future research is expected to continue to make significant contributions to this field.

\section{Methodology}\label{sec3}

Engineering design has grown tremendously in recent years, with increasing demand for more efficient and accurate design processes. To address this challenge, researchers have turned to deep learning techniques to automate the collection of metrics related to mechanical, electrical, and plumbing (MEP) systems in building floor plans. This research project aims to enhance the efficiency of the engineering design process by developing an intelligent system capable of accurately identifying lighting symbols in a floor plan, determining the appropriate type of light, and extracting relevant text to estimate the power requirement of the floor plan. The project utilizes PDF files of 2D floor plans, which are converted into image format for analysis.

\begin{figure*}[t!]%
	\centering
	\includegraphics[width=1\textwidth]{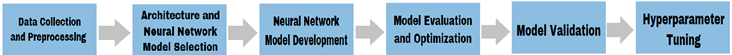}
	\caption{Neural Network Model Development Methodology}\label{fig:fig11}
\end{figure*}

A neural network model is developed using a well-defined methodology to achieve these objectives, as illustrated in Figure~\ref{fig:fig11}. This methodology comprises a series of carefully designed steps, which this section will thoroughly explain.

\subsection{Data Collection and Preprocessing}\label{sec3p1}
\begin{figure*}[ht!]%
	\centering
	\includegraphics[width=1\textwidth]{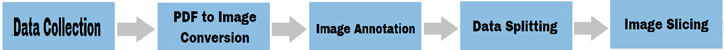}
	\caption{Data Collection and Preprocessing}\label{fig:fig12}
\end{figure*}

Figure~\ref{fig:fig12} presents the steps involve in data collection and preprocessing:

\subsubsection{Data Collection}\label{sec3p1p1}
Developing a neural network model is a complex process that involves several steps, starting with data collection and preprocessing. The process begins with collecting the training data, as no existing dataset was available for the specific problem. In this case, the industry collaborator provided a set of 2D floor plans in PDF format for buildings with different configurations and layouts, which had a vast amount of floor plans.

\subsubsection{PDF to Image Conversion}\label{sec3p1p2}
The next step in the data collection and preprocessing phase was to convert the PDF data to image format. The PDF files were too large to be used directly for training the neural network model. Therefore, it was necessary to convert them to image format. PyMuPDF~\cite{pymupdf2023} was used for the conversion, and the DPI was set to 300 based on trial and experimentation to balance the image size while still maintaining the clarity of the image.

\subsubsection{Image Annotation}\label{sec3p1p3}
\begin{figure*}[h]%
	\centering
	\includegraphics[width=0.75\textwidth]{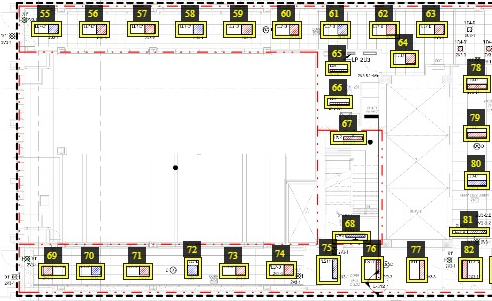}
	\caption{Image Annotation}\label{fig:fig13}
\end{figure*}

The third step in the methodology was manual image annotation using VGG Image Annotator~\cite{dutta2019vgg} (VIA) and converting it to coco data format. The images were manually annotated using the VGG Image Annotator, a popular open-source tool for annotating images. The annotations were made for each image in terms of the location of the lighting symbols and the corresponding text. The annotation data was then converted to the coco data format widely used for object detection tasks. Figure~\ref{fig:fig13} represents a sample image that is annotated as part of the preprocessing steps.

\subsubsection{Data Splitting}\label{sec3p1p4}
The fourth step in the methodology was to split the data into training, validation, and testing datasets. This was done to ensure that the model's performance could be evaluated on unseen data and to avoid over-fitting. This step is crucial in developing a robust and effective neural network model.

\subsubsection{Image Slicing}\label{sec3p1p5}
\begin{figure*}[h]%
	\centering
	\includegraphics[width=.95\textwidth]{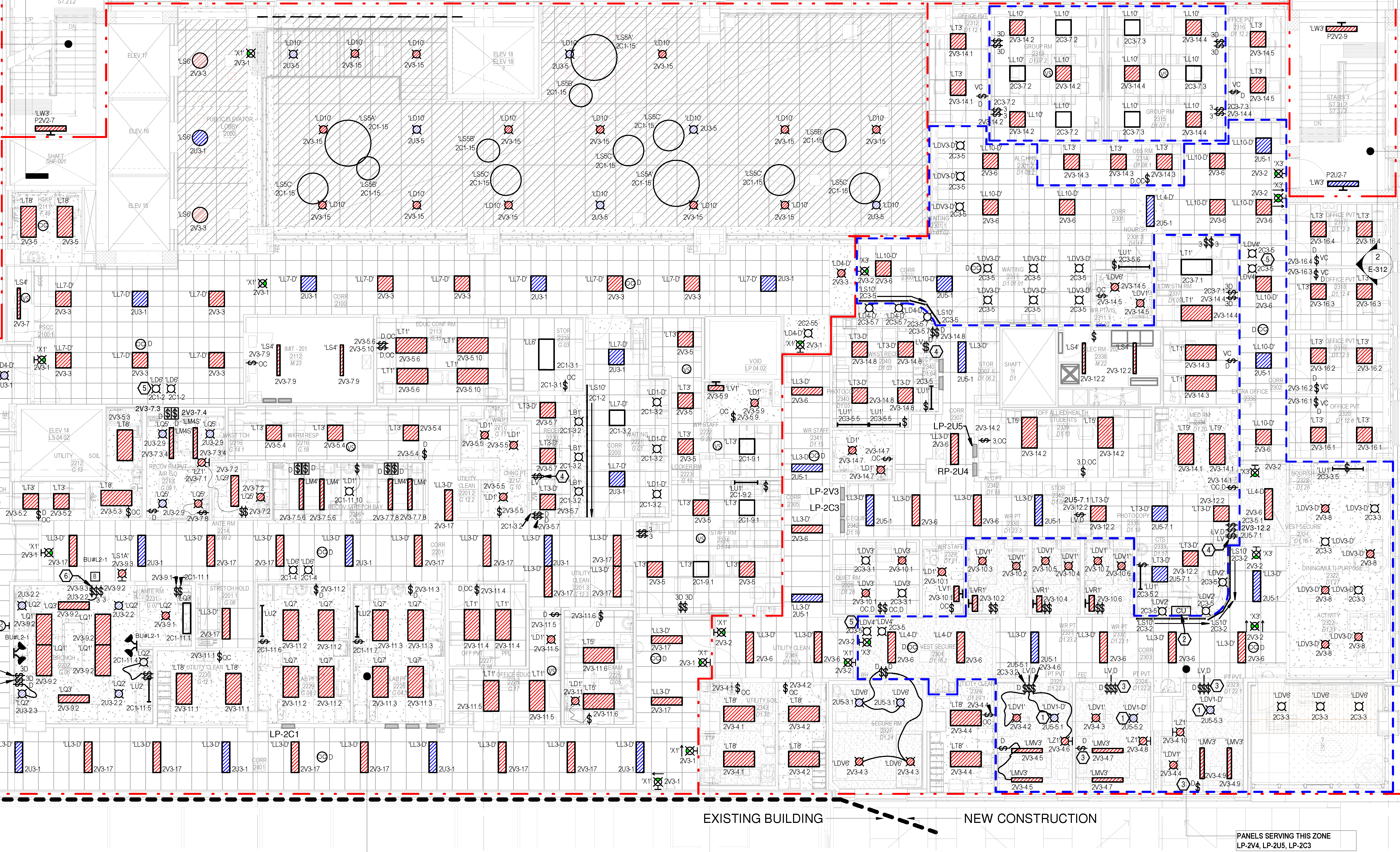}
	\caption{Original Unsliced Image}\label{fig:fig15}
\end{figure*}

\begin{figure}[h]%
	\centering
	\includegraphics[width=.45\textwidth]{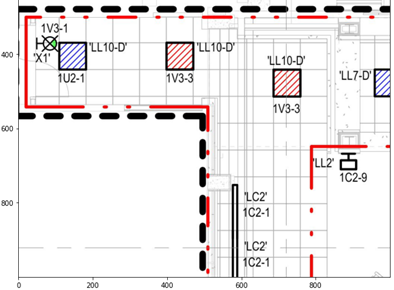}
	\caption{Sliced Component of the Original Image}
	\label{fig:fig16}
\end{figure}

\begin{figure*}[h]%
	\centering
	\includegraphics[width=.87\textwidth]{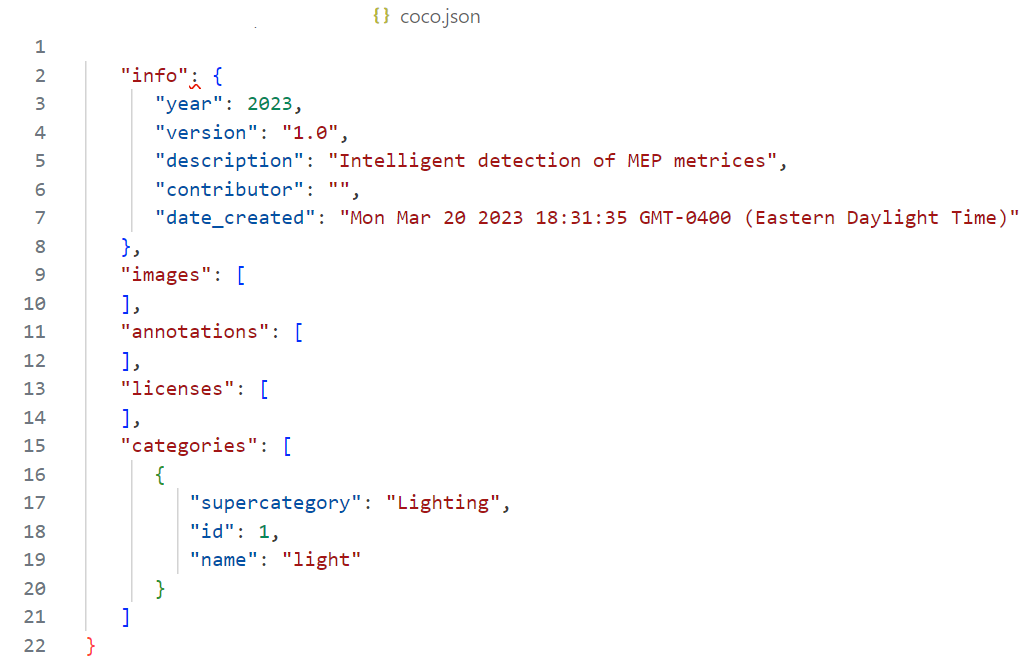}
	\caption{COCO Data Format}
	\label{fig:fig14}
\end{figure*}

Finally, image slicing was performed to slice images into smaller components to reduce training time for the model and improve efficiency. Figures~\ref{fig:fig15} and~\ref{fig:fig16} are showing an original image and it's sliced component respectively. Each image and its coco dataset (see in Figure~\ref{fig:fig14}) were sliced into smaller components. The images were sliced with slice parameters such that each image was sliced into 1000X1000 grids with an overlap ratio of 0.2. This allowed the model to be trained on smaller image components, which reduced the training time and improved efficiency.

\subsection{Architecture and Neural Network Model Selection}\label{sec3p2}

Selecting an appropriate architecture and neural network model is a critical step in developing an effective system for the intelligent detection of MEP metrics. In this study, the Mask R-CNN~\cite{he2017},~\cite{matterport_maskrcnn_2017} algorithm was chosen as the base algorithm for its high accuracy in object detection tasks and ability to detect and segment objects in an image.

Mask R-CNN~\cite{he2017},~\cite{matterport_maskrcnn_2017} is a popular deep-learning algorithm for object detection and image segmentation. It is highly accurate and can create a mask that outlines the exact pixels belonging to each object in an image. This ability to accurately segment images differentiates it from popular computer vision models such as Faster R-CNN~\cite{he2017},~\cite{matterport_maskrcnn_2017}, YOLO~\cite{gu2023systematic}, and SSD~\cite{chandio2022precise}.

\begin{table}
	\centering
	\caption{Object Detection Model Comparison}
	\label{tab:tab1od}
	\begin{tabular}{>{\hspace{0pt}}m{0.15\linewidth}>{\hspace{0pt}}m{0.3\linewidth}>{\hspace{0pt}}m{0.3\linewidth}} 
		\toprule
		\multicolumn{1}{>{\centering\hspace{0pt}}m{0.15\linewidth}}{Model} & \multicolumn{1}{>{\centering\hspace{0pt}}m{0.3\linewidth}}{Advantages}                                                                      & \multicolumn{1}{>{\centering\arraybackslash\hspace{0pt}}m{0.3\linewidth}}{Disadvantages}  \\ 
		\midrule
		
		Mask\par{} R-CNN~\cite{he2017},~\cite{matterport_maskrcnn_2017}       & Object detection and instance segmentation\par{}~in a single model, with high accuracy,\par{}~end-to-end training,~open-source implementation & Computationally expensive, \par{}slower inference speed compared \par{}to other models      \\ \hline
		
		Faster\par{}R-CNN~\cite{gu2023systematic}                     & Good accuracy, faster inference\par{}speed compared to Mask R-CNN   & Requires two\-stage detection, \par{}may miss small objects \\ \hline
		
		YOLO~\cite{gu2023systematic} &Fast inference speed, \par{}real-time object detection           & Lower accuracy compared \par{}to other models,~may miss small objects  \\ \hline
		
		SSD~\cite{chandio2022precise} &Fast inference speed, \par{}real-time object detection, good accuracy for small objects   &Lower accuracy compared \par{}to other models for larger objects   \\ \hline
		                                                     
		\end{tabular}
\end{table}

As shown in Table~\ref{tab:tab1od}, Mask R-CNN offers several advantages over other models, including its ability to perform object detection and instance segmentation in a single model, achieving higher accuracy, and providing more detailed information about the objects in the image. However, it is important to note that Mask R-CNN is computationally expensive and slower than other models, such as Faster R-CNN and YOLO. It is due to its two-stage architecture, which involves more complex computations and higher memory requirements. Thus, for applications where real-time performance is critical, other models like Faster R-CNN and YOLO may be more suitable.

The choice of the model ultimately depends on the application's specific needs. While Mask R-CNN is slower and computationally expensive, its accuracy and segmentation capabilities make it the right choice for this project. In general, there is always a trade-off between accuracy and computational cost when selecting a model. Hence, it is crucial to carefully evaluate each model's pros and cons to determine which best fits the application's requirements.

\subsection{Neural Network Model Development}\label{sec3p3}
The Neural Network Model Development phase aimed to train the Mask R-CNN algorithm to detect and segment MEP metrics. The process involved four stages: 

\subsubsection{Initialization}\label{sec3p3p1}
In this phase, the weights and biases of the neural network are randomly initialized to small values. This step is crucial as it helps avoid getting stuck in local optima.

\subsubsection{Forward Propagation}\label{sec3p3p2}
During forward propagation, the neural network takes input data in the form of images and performs a forward pass through the network, producing an output. In this case, the output is the predicted bounding boxes, objectness scores, and masks for each object in the input image.

\subsubsection{Compute Loss}\label{sec3p3p3}
When training a neural network for object detection, tracking how well the model performs is crucial. One way to do this is by using a loss function to measure the difference between the predicted and true outputs. As the network progresses through each training epoch, a loss value is calculated and printed for each batch. These values and other important information are stored as logs in log files.

%Figure~\ref{fig:fig17} illustrates the progress of training a model for object detection, and the log provides a wealth of valuable information for each epoch and batch. For instance, you can see the epoch number and how many batches are in the epoch, the learning rate, the estimated time remaining, the time taken for each batch, and the amount of memory used during training.

Most importantly, the log includes information on the different components of the object detection task, such as the loss for the region proposal network classification, the loss for bounding box regression, and the accuracy for the classification task. These losses are used to monitor the training progress and adjust the model parameters.

%\begin{figure*}[h]%
%	\centering
%	\includegraphics[width=1\textwidth]{figures/fig17training.png}
%	\caption{Training Progression}\label{fig:fig17}
%\end{figure*}

\subsubsection{Backpropagation}\label{sec3p3p4}
In the backpropagation phase, the loss is backpropagated through the network, and the gradients of the loss with respect to the weights and biases of each neuron are calculated. The specific backpropagation algorithm used in Mask RCNN is stochastic gradient descent with momentum. This algorithm updates the weights and biases of the network based on the gradients calculated during backpropagation. By iteratively adjusting the weights and biases in the direction that minimizes the loss, the network is able to improve its performance over time.

\subsection{Model Evaluation, Optimization, Validation and Hyperparameter Tuning}\label{sec3p4}
This research utilized the validation dataset to evaluate the model's accuracy and optimize its hyperparameters. The primary objective of this process was to improve the performance of the model by adjusting the hyperparameters, including the learning rate, evaluation interval, and number of epochs, among others.

%The primary objective of this process was to improve the performance of the model by adjusting the hyperparameters, including the learning rate, evaluation interval, and number of epochs, among others (Figure~\ref{fig:fig18}).

%\begin{figure*}[ht]%
%	\centering
%	\includegraphics[width=1.0\textwidth]{figures/fig18tuning.png}
%	\caption{Hyperparameter Tuning}\label{fig:fig18}
%\end{figure*}

Hyperparameter tuning is a crucial step in machine learning model building, as it enables us to find the best combination of hyperparameters that results in optimal performance. The hyperparameters were tuned using a trial-and-error process until the desired performance, as measured by Average Precision (AP) and Average Recall (AR), was achieved. 

The model's performance was evaluated using the AP and AR metrics during optimization. These metrics are widely used in object detection tasks and provide a measure of the model's ability to detect and classify objects accurately. The AP metric measures the average precision of the model across a range of intersections over union (IoU) thresholds, while the AR metric measures the average recall of the model across a range of IoU thresholds.

Once the hyperparameters were tuned, the optimized model was tested on a separate test dataset to evaluate its ability to accurately detect lighting symbols. This independent evaluation provided a robust assessment of the model's performance. The rigorous evaluation metrics used during this process ensured that the model's performance was optimal and could generalize to new data. Ultimately, this process led to the development of an intelligent lighting symbol detection model capable of accurately classifying lighting symbols. The final optimized model demonstrated excellent prediction ability, providing high confidence in its ability to be applied in real-world MEP detection scenarios.

\section{Results and Analysis}\label{sec4}
This section presents the research findings and interprets them in the context of the research objectives. The collected data is analyzed comprehensively, and the patterns, relationships, and emerging trends are described. The implications of the findings are also discussed, and key insights that contribute to the understanding of the research problem are highlighted. This section aims to provide a clear and concise presentation of the results and their significance, enabling the reader to evaluate the research outcomes.

\subsection{Analysis of Model Performance During Training}\label{sec4p1}
This subsection provides a comprehensive analysis of the model's performance during training. The data presented include the loss values, accuracy, and memory usage for each epoch, covering a span of 20 epochs. The loss values for each model component are provided, and their trends are analyzed to determine if any overfitting or underfitting occurs. Tensorboard was used as a powerful visualization tool to monitor and analyze the model's performance, providing an interactive dashboard that displays various metrics such as the loss function, accuracy, and learning rate over time.

\begin{figure}[t!]%
	\centering
	\includegraphics[width=1.0\linewidth]{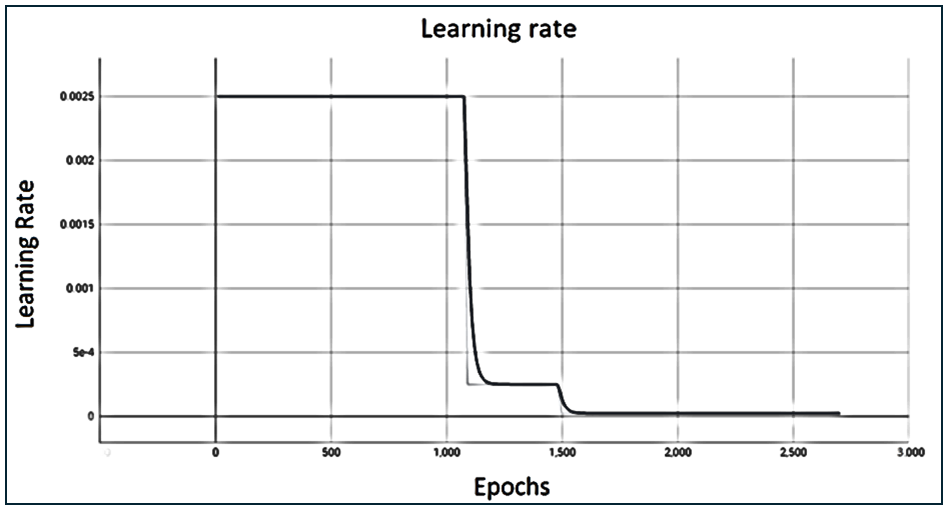}
	\caption{Learning Rate Graph}
	\label{fig:fig19}
\end{figure}

Figure~\ref{fig:fig19} displays the learning rate graph generated by Tensorboard, which depicts the model's learning rate changes over time. The learning rate graph helps determine the appropriate learning rate for optimal performance.

\begin{figure*}[t!]%
	\centering
	\includegraphics[width=.83\textwidth]{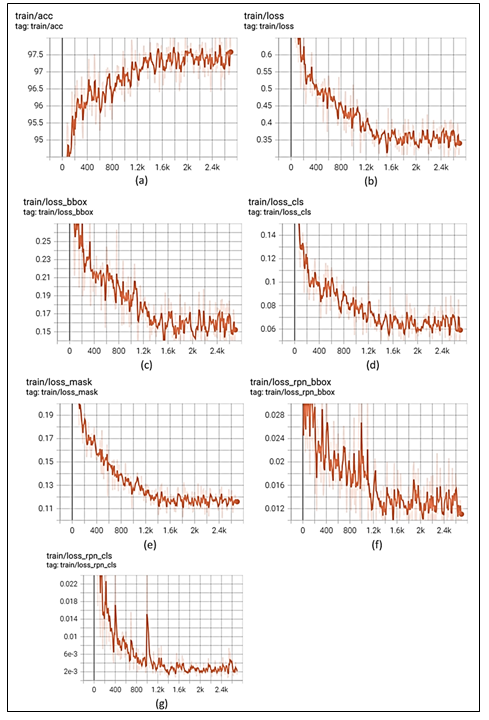}
	\caption{Training Accuracy and Loss Graphs}
	\label{fig:accu}
\end{figure*}

In addition to the learning rate graph, the Figure~\ref{fig:accu}(a) presents the accuracy of the model and Figures~\ref{fig:accu}(b) to (g) are showing the training loss graphs generated by Tensorboard. These graphs show how the model's loss changes over time during training. The figures show that the training loss decreases steadily over time, indicating that the model effectively learns from the data and improves its performance. 

\begin{table*}[]
	\caption{Training Metrics}
	\label{tab:tf1}
	\setlength{\tabcolsep}{3pt}
	%\centering
	%\resizebox{\columnwidth}{!}
	{%
		\begin{tabular}{|c|c|c|c|c|c|c|c|c|c|c|}
			\hline
			Epoch &
			\begin{tabular}[c]{@{}c@{}}Learning\\  Rate\end{tabular} &
			\begin{tabular}[c]{@{}c@{}}Training \\ Time\end{tabular} &
			Memory &
			Loss\_rpn\_cls &
			Loss\_rpn\_bbox &
			Loss\_cls &
			Accuracy &
			Loss\_bbox &
			Loss\_mask &
			Loss \\ \hline
			1  & 2.500e-03 & 0:40:42 & 2543 & 0.1327 & 0.0245 & 0.3299 & 86.1719 & 0.2994 & 0.5616 & 1.3480 \\ \hline
			10 & 2.500e-04 & 0:10:28 & 2904 & 0.0019 & 0.0125 & 0.0716 & 97.0508 & 0.1685 & 0.1259 & 0.3804 \\ \hline
			20 & 2.500e-05 & 0:00:02 & 2904 & 0.0014 & 0.0105 & 0.0570 & 97.5586 & 0.1559 & 0.1161 & 0.3409 \\ \hline
	\end{tabular}
	}
\end{table*}

Table~\ref{tab:tf1} shows that the loss\_rpn\_cls, loss\_rpn\_bbox, loss\_cls, and loss\_bbox decrease significantly with each epoch, while the accuracy improves. The loss\_mask, however, shows a slight decrease in the first epoch but stays relatively stable in the following epochs. The results indicate that the algorithm used in this research study is effective in detecting MEP metrics.

\subsection{Performance Evaluation}
The performance of the proposed intelligent detection model for MEP metrics was evaluated using the mean Average Precision (mAP) and mean Average Recall (mAR) metrics. The results of the evaluation show that the model achieved a mAP of 0.760 and a mAR of 0.810 for bounding box annotations with an IoU threshold of 0.50:0.95, and a maximum detection of 100. 

%The\textbf{ results} of the evaluation show that the model achieved a mAP of 0.760 and a mAR of 0.810 for bounding box annotations with an IoU threshold of 0.50:0.95, and a maximum detection of 100 (Figure~\ref{fig:fig22}). 

%\begin{figure*}[h]%
%	\centering
%	\includegraphics[width=1\textwidth]{figures/fig22bboxresults.png}
%	\caption{bbox Evaluation Results}\label{fig:fig22}
%\end{figure*}

\begin{figure*}[h]%
	\centering
	\includegraphics[width=.9\textwidth]{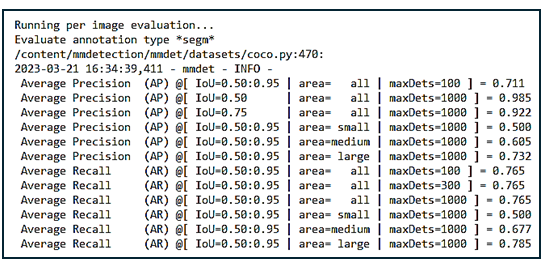}
	\caption{Segmentation Evaluation Results}\label{fig:fig23}
\end{figure*}

For segmentation annotations, the model achieved a mAP of 0.711 and a mAR of 0.765 with the same IoU threshold and maximum detection (Figure~\ref{fig:fig23}). The model also achieved high accuracy for different object sizes, with a mAP of 0.700 for small objects, 0.662 for medium-sized objects, and 0.780 for large objects. The mAR results for different object sizes were 0.700, 0.725, and 0.830, respectively. The results demonstrate the effectiveness of the proposed model in accurately detecting MEP metrics.

\subsection{Evaluation Results}
\begin{figure*}[h]%
	\centering
	\includegraphics[width=0.9\textwidth]{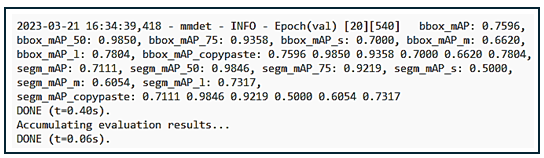}
	\caption{Evaluation Results for Epoch 20}\label{fig:fig24}
\end{figure*}

The following subsection provides an analysis of the results obtained from the evaluation of the performance of MEP metrics detection. The results are presented in terms of detection accuracy metrics such as bbox\_mAP, segm\_mAP, and their variations. The evaluation was conducted using the MMDetection framework and the evaluation results are shown in Figure~\ref{fig:fig24}.

The evaluation metrics provide insight into the performance of the model in terms of its ability to detect MEP metrics accurately. The bbox\_mAP metric measures the mean average precision of bounding box detection, while segm\_mAP measures the mean average precision of segmentation. The variations of these metrics, such as bbox\_mAP\_50 and segm\_mAP\_75, measure the average precision at different intersection over union (IoU) thresholds.

The results obtained from the evaluation demonstrate that the model has achieved high accuracy in detecting MEP metrics, with bbox\_mAP and segm\_mAP values of 0.7596 and 0.7111, respectively. The results also show that the model performs well at different IoU thresholds, with bbox\_mAP\_50 and segm\_mAP\_75 values of 0.9850 and 0.9219, respectively. The analysis of the results provides insight into the performance of the model and highlights areas for further improvement.

\subsection{Detection and Extraction of Lighting Symbols for Energy Consumption Prediction}\label{sec4p4}
\begin{figure}[t!]%
	\centering
	\includegraphics[width=0.45\textwidth]{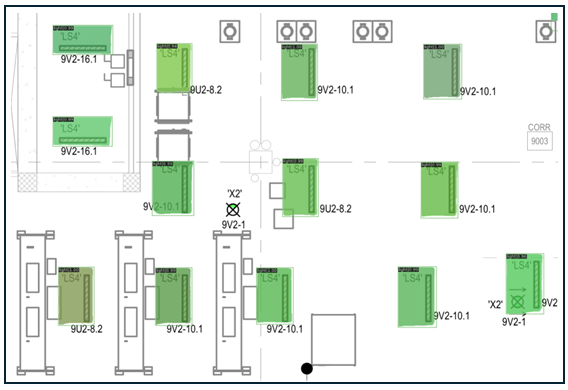}
	\caption{Detected Lighting Symbols in Test Image}
	\label{fig:fig25}
\end{figure}

The trained model was applied to an unseen image to evaluate its performance. Figure~\ref{fig:fig25} displays the output of the model, which successfully identified the lighting symbols in the image. 

\begin{figure}[t!]%
	\centering
	\includegraphics[width=.13\textwidth]{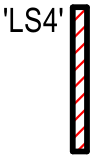}
	\caption{Extraction of Lighting Symbols and Associated Text}
	\label{fig:fig26}
\end{figure}

The lighting symbols were then extracted, and using the Pytesseract tool, the text associated with each symbol was extracted shown in Figure~\ref{fig:fig26}.

This text provides information about the type of light present in the floor plan, which can be used to predict the overall power consumption. The successful extraction of text from the image demonstrates the effectiveness of the developed model in accurately identifying and extracting MEP metrics from floor plan images.

\section{Conclusion}\label{sec5}
In this research work, we have applied the Mask Region-based Convolutional Neural Network (Mask RCNN) to intelligently detect MEP symbols from floor plans to improve the engineering design process’s efficiency, accuracy, and cost-effectiveness.

The study achieved bbox\_mAP and segm\_mAP values of 0.7596 and 0.7111, respectively, and performed well at different IoU thresholds, with bbox\_mAP\_50 and segm\_mAP\_75 values of 0.9850 and 0.9219, respectively.  These results represent a significant improvement over the traditional methods.

The developed approach can automate the manual work involved in MEP symbol detection, reducing design time and increasing workflow efficiency.  It has the potential to significantly impact the fields of architecture, construction, and real estate by improving energy efficiency during the design process and contributing to the net-zero greenhouse gas emissions goal the Government of Canada set by 2050.

The method employed for intelligently detecting MEP metrics represents a substantial advancement in the MEP engineering sector. It establishes a groundwork for future studies aimed at enhancing its precision and versatility across various image types and qualities.

\balance
\bibliographystyle{IEEEtran}
\bibliography{sn-bibliography}

\clearpage
%\newpage

\begin{IEEEbiography}[{\includegraphics[width=1in,height=1.25in,clip,keepaspectratio]{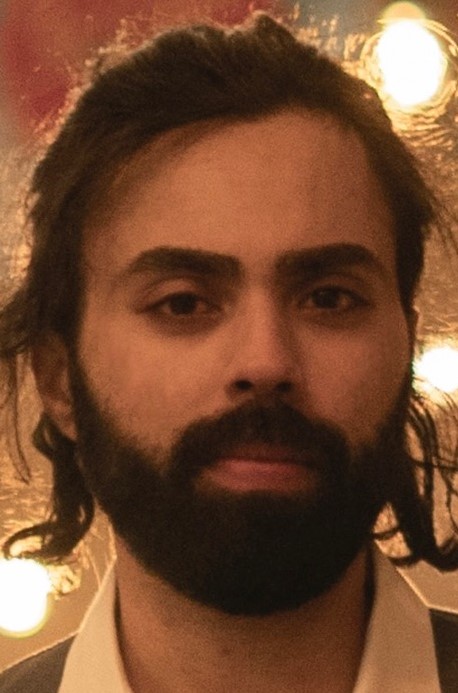}}]{Tarandeep S. Mandhiratta} Tarandeep S. Mandhiratta received his Master of Applied Computing degree from Wilfrid Laurier University in 2023 and a Bachelor of Engineering in Computer Engineering from Savitribai Phule Pune University in 2020. 

Tarandeep has experience in software engineering, data analysis, artificial intelligence, machine learning, and deep learning. He has applied his expertise in developing innovative solutions using advanced algorithms, neural networks, and data-driven methodologies. 
\end{IEEEbiography}

\begin{IEEEbiography}[{\includegraphics[width=1in,height=1.25in,clip,keepaspectratio]{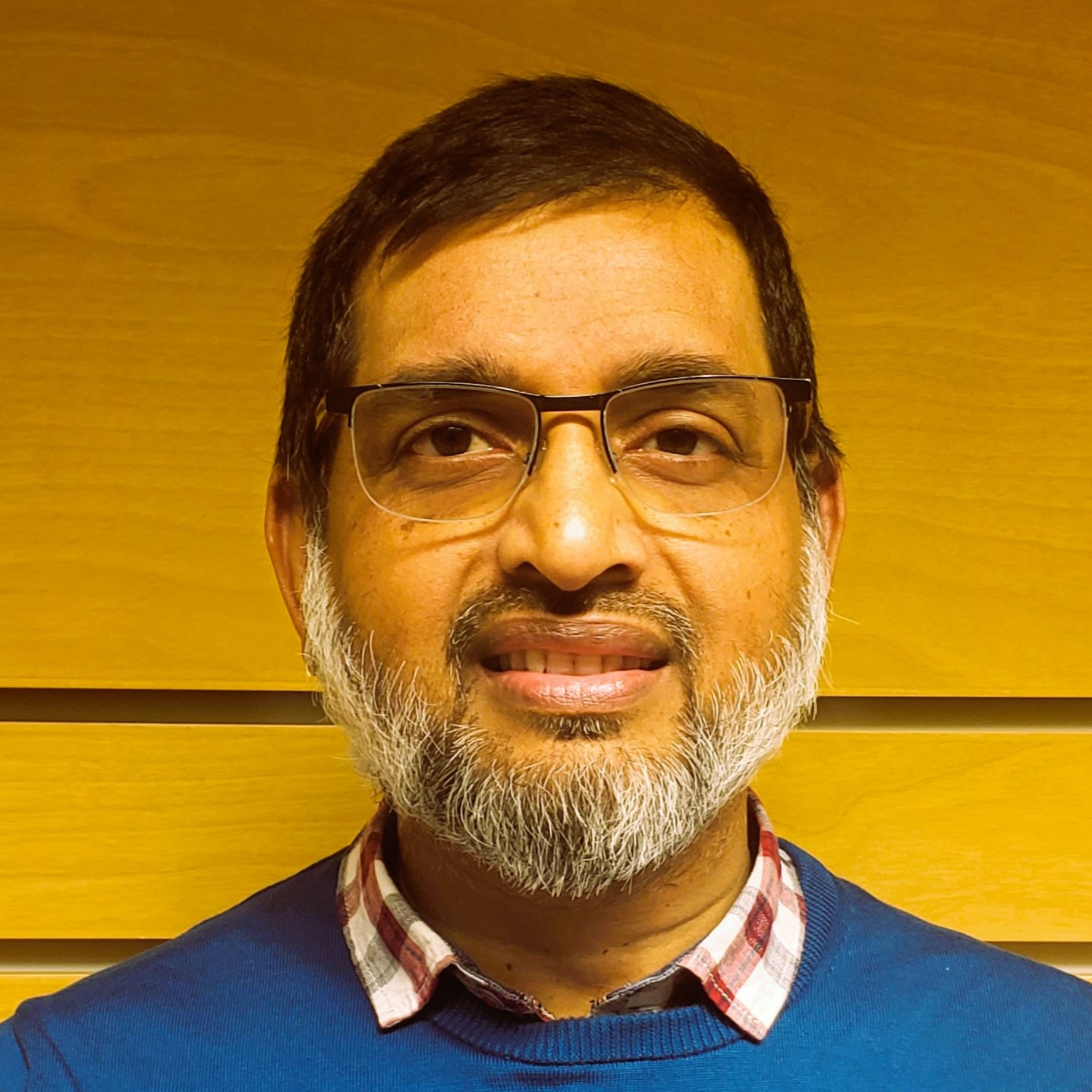}}]{ANK Zaman} obtained his PhD and MSc in Computer Science from the School of Computer Science, University of Guelph, and the Computer Science Program at the University of Northern British Columbia (UNBC), Prince George, BC, Canada. His research interests encompass Machine Learning, Cybersecurity, Data Science, and Computer Vision. His work aims to develop robust machine learning algorithms for threat detection, data analytics, and computer vision applications, focusing on enhancing digital systems' security and efficiency and harnessing visual data's potential for various domains.  He is currently a faculty member at the Dept. of Physics \& Computer Science, Wilfrid Laurier University (WLU), Waterloo, Ontario, Canada. 
\par
Dr. Zaman has been volunteering for the IEEE Kitchener-Waterloo (KW) (Region 7, Canada) 
2013, and he is the Chair of the IEEE Computer Society, IEEE KW, Canada. He is also an active member at  Canadian Artificial Intelligence Association (CAIAC).
\end{IEEEbiography}

\begin{IEEEbiography}[{\includegraphics[width=1in,height=1.25in, clip, keepaspectratio]{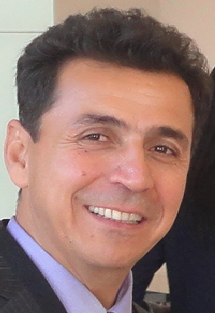}}]{Abdul-Rahman Mawlood-Yunis} Abdul-Rahman Mawlood-Yunis is an assistant professor at Wilfrid Laurier University and the author of the ook ``Android for Java Programmers.'' 
My expertise lies in Mobile Apps and Smart Devices, Software Engineering, Natural Language Processing (NLP), Distributed Computing, and, Algorithm Design. My recent research projects have centred around Machine Learning on mobile devices and NLP applications for languages with resource constraints.
	%and all authors may include
%biographies. Biographies are often not included in conference-related
%papers. This author became a Member (M) of IEEE in 1976, a Senior
%Member (SM) in 1981, and a Fellow (F) in 1987. The first paragraph may
%contain a place and/or date of birth (list place, then date). Next,
%the author's educational background is listed. The degrees should be
%listed with type of degree in what field, which institution, city,
%state, and country, and year the degree was earned. The author's major
%field of study should be lower-cased.
%
%The second paragraph uses the pronoun of the person (he or she) and not the
%author's last name. It lists military and work experience, including summer
%and fellowship jobs. Job titles are capitalized. The current job must have a
%location; previous positions may be listed
%without one. Information concerning previous publications may be included.
%Try not to list more than three books or published articles. The format for
%listing publishers of a book within the biography is: title of book
%(publisher name, year) similar to a reference. Current and previous research
%interests end the paragraph.
%
%The third paragraph begins with the author's
%title and last name (e.g., Dr.\ Smith, Prof.\ Jones, Mr.\ Kajor, Ms.\ Hunter).
%List any memberships in professional societies other than the IEEE. Finally,
%list any awards and work for IEEE committees and publications. If a
%photograph is provided, it should be of good quality, and
%professional-looking. Following are two examples of an author's biography.
\end{IEEEbiography}

%	[{\includegraphics[width=1in,height=1.25in,clip,keepaspectratio]{author3.png}}]{Third C. Author, Jr.} (M'87) received the B.S. degree in mechanical
%engineering from National Chung Cheng University, Chiayi, Taiwan, in 2004
%and the M.S. degree in mechanical engineering from National Tsing Hua
%University, Hsinchu, Taiwan, in 2006. He is currently pursuing the Ph.D.
%degree in mechanical engineering at Texas A{\&}M University, College
%Station, TX, USA.
%
%From 2008 to 2009, he was a Research Assistant with the Institute of
%Physics, Academia Sinica, Tapei, Taiwan. His research interest includes the
%development of surface processing and biological/medical treatment
%techniques using nonthermal atmospheric pressure plasmas, fundamental study
%of plasma sources, and fabrication of micro- or nanostructured surfaces.
%
%Mr. Author's awards and honors include the Frew Fellowship (Australian
%Academy of Science), the I. I. Rabi Prize (APS), the European Frequency and
%Time Forum Award, the Carl Zeiss Research Award, the William F. Meggers
%Award and the Adolph Lomb Medal (OSA).

\EOD

\end{document}